\documentclass[letterpaper, 10pt, conference]{ieeeconf}
\IEEEoverridecommandlockouts
\usepackage[utf8]{inputenc}
\usepackage[T1]{fontenc}
\usepackage{graphicx}
\usepackage{amsmath,amssymb,amsfonts}
\usepackage{textcomp}
\usepackage{xcolor}
\usepackage{cite}
\usepackage{url}
\usepackage{siunitx}
\usepackage{booktabs}
\usepackage{multirow}
\usepackage{comment}
\usepackage{tikz}
\usetikzlibrary{shapes.geometric,arrows.meta,positioning,fit,backgrounds}

\graphicspath{{figures/}}

\begin{document}

\title{From Sky to Soil: A Morphing
  Aerial-Ground Robot for Seed Deployment}

\author{Namai~Chandra$^{1}$ and Lining~Yao$^{2}$%
  \thanks{$^{1}$N.~Chandra is with the Department of Electrical Engineering,
    Indian Institute of Technology Madras, Chennai 600036, India
    (e-mail: \texttt{23f3000200@es.study.iitm.ac.in}).}%
  \thanks{$^{2}$L.~Yao is with the Morphing Matter Lab,
    Department of Mechanical Engineering, University of California,
    Berkeley, CA 94720 USA
    (e-mail: \texttt{liningy@berkeley.edu}).}%
}

\maketitle
\thispagestyle{empty}
\pagestyle{empty}

\begin{abstract}
Aerial seed broadcasting can reach remote restoration sites, but provides
limited control over seed placement within the soil. This paper presents
a geometry-assisted, tri-functional morphing robot that combines aerial
access, ground locomotion, and controlled-depth seed embedding in a
\emph{fly-drive-plant} architecture. After landing, the platform
reconfigures into a four-wheeled planting configuration: an electronically
coupled dual-motor drive folds the rear arms outward to form ground
wheels, while a descending front tray engages the propulsion motors with
a drill gear train. Reusing the propulsion motors for drilling eliminates
a dedicated drill drive. The planting sequence forms a hole, dispenses a
seed, and allows the vehicle to reposition on the ground or return to its
flight configuration. Ground mobility supports repeated planting without
requiring a separate flight between adjacent sites. A companion
controller issues reconfiguration, tray, and seed-gate commands, while a
dedicated autopilot handles flight control. The morphing and planting mechanisms are validated
using a hardware prototype that demonstrates ground repositioning and
seed embedding. This proof of concept establishes a hardware basis
for aerial-ground seed embedding through coordinated reconfiguration and
actuator reuse.
\end{abstract}

\begin{keywords}
Aerial systems: mechanics and control, mechanism design, agricultural
automation, field robots, reconfigurable robots.
\end{keywords}

\section{Introduction}
\label{sec:intro}

Reforestation, post-fire revegetation, and ecological
restoration are increasingly bottlenecked by access. The slopes, burn
scars, and remote watersheds where restoration effort is most needed are
precisely the terrain where humans, tractors, and wheeled planters are
least deployable~\cite{zahawi2015uavforest}. Unmanned aerial vehicles
have become the instrument of choice for reaching this terrain, and a
growing body of work uses them to deliver seed from the air at
scale~\cite{stamatopoulos2024uavreview,kulkarni2023uavreforestation}.
Aerial access alone does not ensure seedling establishment.
Broadcasting and seed-ball drops leave seeds on the surface, exposed to
desiccation and seed predators~\cite{luo2023selfburying,joyce2024predation}.
In field trials at three degraded rainforest sites in Australia, Doust
\textit{et al.}\ reported eight-month seedling establishment of
\SIrange{1.4}{3.3}{\percent} of viable seeds sown in surface-broadcast
treatments, compared with \SIrange{22.6}{33.4}{\percent} in shallow-burial
treatments~\cite{doust2006directseeding}. Establishment here measures
surviving seedlings relative to viable seeds sown, combining germination
and subsequent survival. The authors identified desiccation and predation
as likely causes of poor surface establishment.
These results motivate control over seed placement within the soil,
with sowing depth selected for the species and site conditions.
\begin{figure}[t]
  \centering
  \includegraphics[width=0.7\columnwidth]{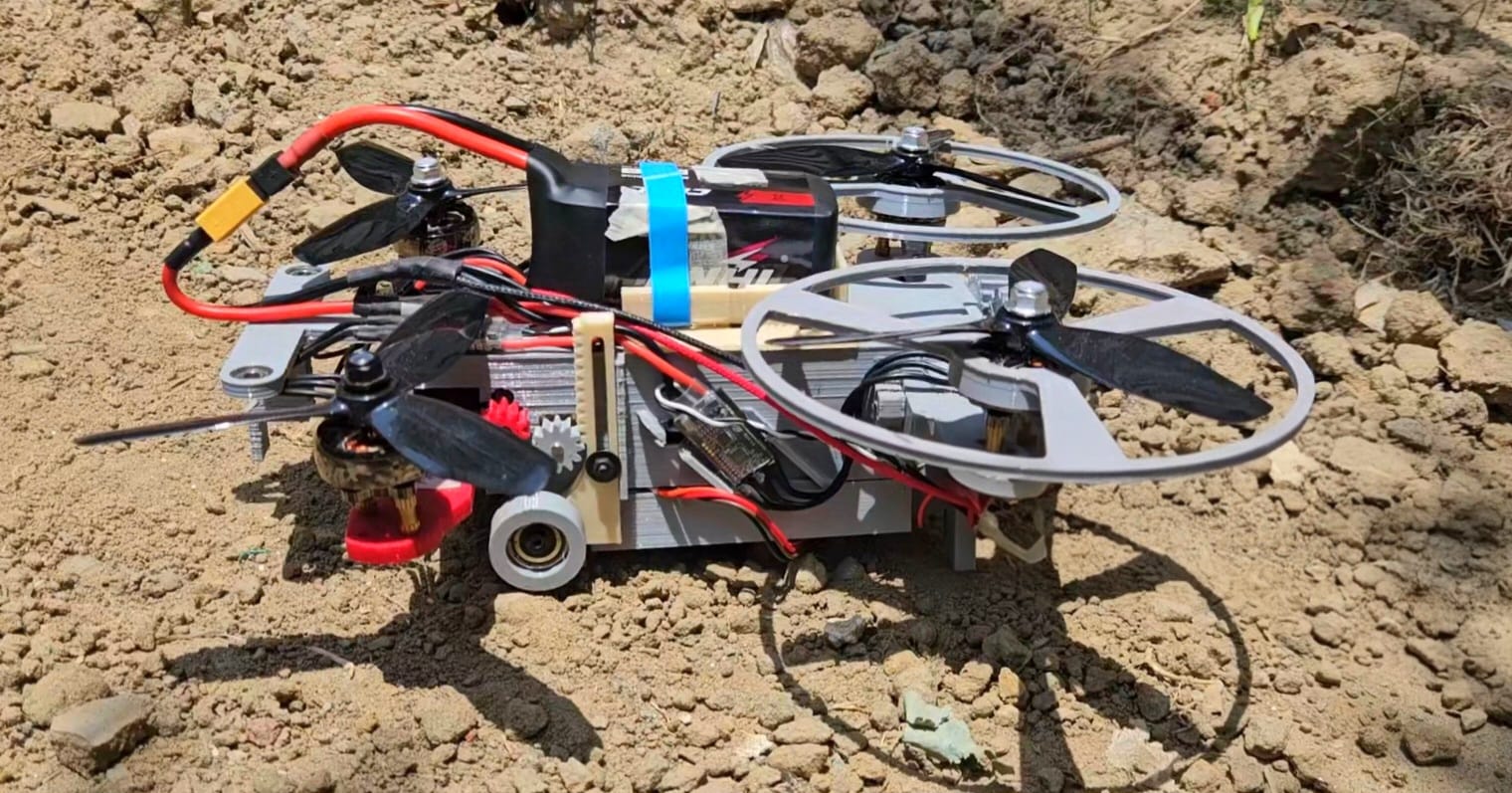}\\[1pt]
  \small (1) Flight configuration\\[4pt]
  \includegraphics[width=0.7\columnwidth]{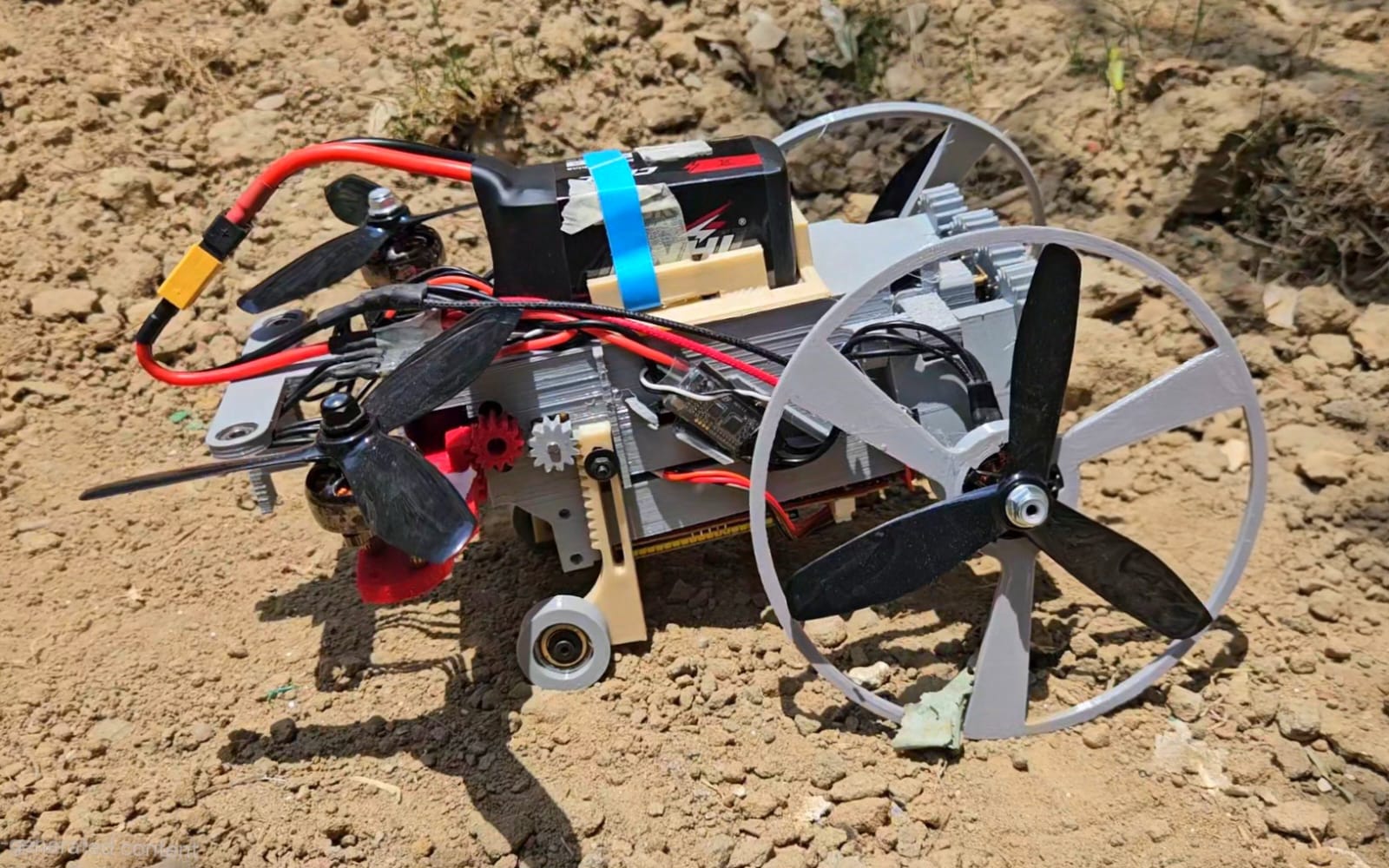}\\[1pt]
  \small (5) Plant (morphed) configuration
  \caption{Prototype on terrain in (1)~flight configuration and
    (5)~plant (morphed) configuration, corresponding to the numbered
    CAD stages of Fig.~\ref{fig:configs}.}
  \label{fig:live_photos}
\end{figure}
\begin{figure*}[!t]
  \centering
  \includegraphics[width=\textwidth]{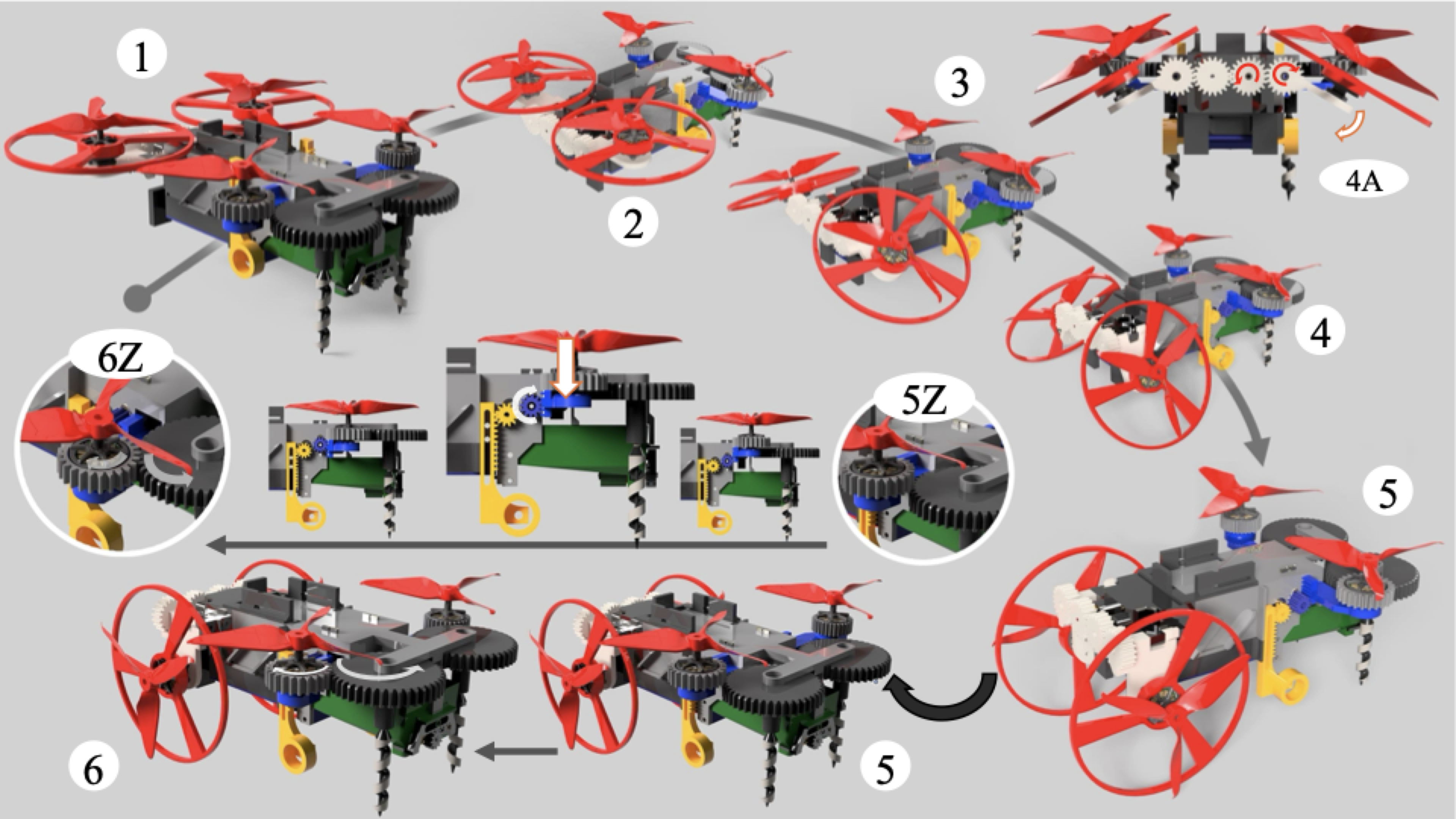}
  \caption{Complete morphing sequence from flight to plant configuration.
    \textbf{Stage~1}: drone (flight) mode; all four arms extended,
    propeller-guard rings upright.
    \textbf{Stage~2}: rear arms begin rotating outward
    ($\theta \approx 0^{\circ}$ from flight position).
    \textbf{Stage~3}: rear arms at $\approx 30^{\circ}$.
    \textbf{Stage~4}: rear arms at $\approx 60^{\circ}$;
    \textbf{4A}~(inset, top-right) shows the back view of the
    gear train responsible for converting rack travel into rear-arm
    rotation (the mechanism by which propeller-guarded arms fold
    outward to become ground wheels).
    \textbf{Stage~5}: rear arms at $90^{\circ}$, plant mode
    reached; a second view of Stage~5 (rotated perspective) shows the
    platform grounded on all four wheels.
    \textbf{5Z}~(zoom inset): front-tray gear disengaged: the
    brushless motor shafts and the drill-drive gear are not yet meshed.
    Between \textbf{5Z} and \textbf{6Z}, three intermittent frames
    (read right to left) show the front tray descending via a
    rack-and-pinion mechanism: the pinion rotates clockwise,
    translating the front motor tray downward.
    \textbf{6Z}~(zoom inset): front tray fully descended: the
    brushless motor shafts now mesh with the drill-drive gear,
    engaging the drill bits.
    \textbf{Stage~6}: fully deployed plant configuration with drill
    bits engaged and ready for seed embedding.}
  \label{fig:configs}
\end{figure*}
Aerial vehicles can bypass terrain obstacles, but flight energy limits
mission duration; ground vehicles can travel more efficiently, but
slopes, surface undulations, and obstacles can prevent travel from the
deployment point to the planting site and back~\cite{kalantari2014hytaq,sihite2023m4}.
Morphing and aerial-ground hybrid multirotors address these mobility
constraints by squeezing through gaps~\cite{falanga2019foldable},
rolling along the ground~\cite{kalantari2013hytaq,kalantari2014hytaq}, or
repurposing propellers as wheels~\cite{sihite2023m4}. Drone-based seeders
provide aerial seed delivery~\cite{lee2025uavseed,kulkarni2023uavreforestation},
but precise release improves where a seed lands, not whether it is
covered by soil.
Combining aerial access, ground repositioning, and controlled-depth seed
embedding therefore requires integrating mobility with soil interaction.
\textbf{This paper presents a \emph{fly-drive-plant}
platform that integrates aerial access, ground repositioning, and
controlled-depth seed embedding through reconfiguration and
propulsion-motor reuse, all valideated with a proof-of-concept.} Autonomous navigation is outside the present
evaluation: the prototype uses ArduPilot and Mission Planner for flight,
while mechanism commands are issued through a companion controller
(Section~\ref{sec:control}). Bench morphing tests, inertial measurements,
and outdoor seed-placement trials evaluate the resulting hardware.

The vehicle flies to a target location, lands, and morphs into a
four-wheeled planting configuration: the rear arms fold outward as ground
wheels, and the front motor tray descends to couple the propulsion motors
to a drill gear train. The drill opens a hole at a controlled and a configurable depth of up
to \SI{30}{\milli\metre}, a seed is dispensed into
it, and the vehicle either advances on its deployed wheels to the next
site or morphs back to flight. The prototype thereby combines the aerial
reach of a drone, the ground mobility of a rover, and the soil-engaging
action of a planting machine in an airframe of less than
\SI{500}{\gram} (battery excluded). The contributions of this paper are: (i)~a tri-functional morphing
architecture that reuses the propeller guards as ground wheels and the
propulsion motors as the drill drive; (ii)~a geometry-assisted,
open-loop morph command in which mechanical end-stops and
non-backdrivable gearing define and hold the terminal configurations
without arm-position sensing; and (iii)~a hardware evaluation comprising
repeated bench morphing cycles, inertial characterization of both
transitions, and outdoor drill-and-dispense trials.

\section{Related Work}
\label{sec:related}

\textbf{Morphing UAVs.}~Reconfigurable multirotors have attracted
significant attention as a means of extending the operating envelope of
the rigid frame. Falanga \textit{et al.}\ introduced the foldable drone,
contracting its arms in flight to squeeze through narrow apertures and
retune control allocation to the time-varying
inertia~\cite{falanga2019foldable}. Fabris \textit{et al.}\ extended
this with a geometry-aware compensation scheme for aerodynamic effects
caused by overlap between propellers and the central body during
morphing~\cite{fabris2020geometry}, and later demonstrated an autonomous
morphing drone that folds its arms to traverse narrow passageways in
flight~\cite{fabris2022crash2squash}. Wang
\textit{et al.}\ proposed a Sarrus-linkage frame for symmetric
expansion and contraction~\cite{wang2024sarrus}, and Yeh \textit{et
  al.}\ demonstrated a central-servo, gear-and-rack architecture for
vertically foldable arms with embedded grasping
fingers~\cite{yeh2025verticallyfoldable}. Recent reviews catalog a
broader design space and consistently identify mechanical and energetic
overhead as the dominant cost of
morphology~\cite{xing2024morphingreview,acar2025morphingmechanisms,derrouaoui2023morphingcontrol}.
These systems use reconfiguration for flight adaptation and, in some
cases, object grasping~\cite{falanga2019foldable,yeh2025verticallyfoldable};
the cited demonstrations do not address controlled seed depth.

\textbf{Aerial-Ground Hybrid Platforms.}~Kalantari and Spenko's HyTAQ
pioneered the dual-mode aerial-terrestrial platform, surrounding the
airframe with a passively rolling cage to amortize flight energy by
rolling on the ground~\cite{kalantari2013hytaq,kalantari2014hytaq}.
Atay \textit{et al.}\ formalized control allocation for bimodal
rotary-wing rolling-flying vehicles~\cite{atay2021bimodal}. Sihite
\textit{et al.}\ demonstrated the M4 Morphobot, which repurposes
shrouded propeller assemblies for flight, wheeled and legged locomotion,
and object grasping~\cite{sihite2023m4}, and Zhang \textit{et al.}\ developed an
autonomous tilting multirotor that redirects thrust to roll across rough
terrain~\cite{zhang2023aqthr}. These systems demonstrate ground mobility
and, in M4, object manipulation; their reported tasks do not include
soil drilling or seed embedding.

\textbf{Drone-Based Seeding and Planting.}~UAVs have become standard
tools for monitoring restoration sites~\cite{zahawi2015uavforest} and,
more recently, for active seed delivery. Lee \textit{et al.}\ presented
a UAV-based precision seed-dropping system using biochar-coated seed
balls released individually from low altitude~\cite{lee2025uavseed}.
Such systems control release location but do not actively set seed
burial depth. Individual placement may suit costly seeds, whose prices
vary substantially among species~\cite{raupp2020seedcost}. Earlier patent
concepts proposed aerial platforms for combined crop dusting, planting,
and fertilizing~\cite{burema2016aerialfarm}, without a reconfigurable
ground mode or controlled embedding depth.

\textbf{Seed Placement and Seedling Survival.}~Surface-sown seeds
can fail to establish through desiccation, predation, and seedling
mortality~\cite{doust2006directseeding,joyce2024predation}. Field
comparisons show that shallow burial can improve establishment relative
to surface broadcasting, with outcomes depending on seed size and
site conditions~\cite{doust2006directseeding}. Species selection further
governs direct-seeding outcomes: Naruangsri \textit{et al.}\ identify
traits that predict which tree species establish well from
directly sown seed in degraded tropical forest~\cite{naruangsri2024directseeding}.
Luo \textit{et al.}\ attack this from the seed side, designing
self-burying carriers that drill into the ground after
landing~\cite{luo2023selfburying}; the present work attacks it from the
platform side, bringing a powered drill to the seed.

\textbf{Actuation Minimization.}~Every independent drive adds mass,
electrical load, and a failure mode, motivating designs that achieve
large shape change with few actuators: compliant and origami-inspired
structures~\cite{chen2021origamispring,kim2022dpneunet}, underactuated
hands~\cite{lee2020twister}, and minimalist
brachiators~\cite{javadi2023acromonk} all show a single motor can
coordinate many joints through kinematic coupling. These studies motivate
actuator reuse in the present platform, where rear-arm reconfiguration,
ground-wheel deployment, and drill engagement are integrated within a
multirotor airframe.

Table~\ref{tab:comparison} compares the platform with representative
morphing, aerial-ground, and drone-seeding systems, including their
physical tasks and seed-placement methods.

\begin{table*}[!t]
  \centering
  \caption{Comparison with Representative Hybrid and Seeding Platforms}
  \label{tab:comparison}
  \scriptsize
  \setlength{\tabcolsep}{1.25pt}
  \renewcommand{\arraystretch}{0.88}
  \begin{tabular}{lccccccccc}
    \toprule
    System & Mass (g) & Modes & \shortstack{Morph\\drive} &
    \shortstack{Morph\\DoF} & \shortstack{Actuator\\reuse} &
    \shortstack{Ground\\contact} & Task & \shortstack{Seed\\placement} &
    \shortstack{Depth\\set} \\
    \midrule
    Foldable Drone~\cite{falanga2019foldable} & 580 & flight & 4 servos & 4 & no & NA & grasp & NA & NA \\
    M4~\cite{sihite2023m4} & 5600\textsuperscript{a} & air/wheel/leg & 8 servos & 8 hip & prop./wheel & wheels/legs & grasp & NA & NA \\
    Geometry-morphing quad~\cite{fabris2020geometry} & NR & flight & arm servos & 4 & no & NA & NA & NA & NA \\
    HyTAQ~\cite{kalantari2013hytaq} & NR & air/roll & passive cage & 0 & prop./roll & cage & mobility & NA & NA \\
    Sarrus frame~\cite{wang2024sarrus} & NR & flight & 1 actuator & 1 & no & NA & NA & NA & NA \\
    Vert.\ foldable UAV~\cite{yeh2025verticallyfoldable} & 805 & flight & 1 servo & 1 coupled & no & NA & grasp & NA & NA \\
    Seed-drop UAV~\cite{lee2025uavseed} & NR & flight & fixed frame & 0 & no & none & deliver & surface & no \\
    Reforestation UAV~\cite{kulkarni2023uavreforestation} & NR & flight & fixed frame & 0 & no & none & deliver & aerial drop & no \\
    \textbf{Ours} & \textbf{$\sim$800}\textsuperscript{b} & \textbf{air/drive} & \textbf{2 DC + 2 servos} & \textbf{3} & \textbf{prop./drill} & \textbf{guards/wheels} & \textbf{plant} & \textbf{bored hole} & \textbf{$\leq$30\,mm} \\
    \bottomrule
  \end{tabular}
  \par\smallskip
  \parbox{\textwidth}{\footnotesize
    \textsuperscript{a}Excluding stereo camera.
    \textsuperscript{b}Prototype all-up mass with the 4S battery.
    Morph DoF exclude rotor and wheel spin; our three coordinates are
    the two rear-arm angles and front-tray displacement.
    NR: not reported; NA: not applicable.}
\end{table*}

\section{Problem Formulation and Design Requirements}
\label{sec:problem}

\textbf{Planting task.} Consider a restoration site that must be reached
by air and contains several planting locations a short distance apart
along a row. At each location, the robot must form a hole in unprepared
ground, place one seed of diameter at most $d_s$ into it, and then either
travel to the next location or take off. The robot must therefore apply
a tool to the soil and remain stationary while it does so.
The shallow-burial treatments that improved establishment
in~\cite{doust2006directseeding} placed seed approximately
\SIrange{5}{20}{\milli\metre} deep, which sets the order of the required
tool reach; the appropriate depth remains species- and site-dependent.
Because adjacent locations are close, a takeoff and landing at each one
would spend flight energy on short hops, motivating ground repositioning
(Section~\ref{sec:performance}).

\textbf{Research question.} Can a single multirotor airframe provide
aerial access, ground repositioning between nearby locations, and
controlled-depth soil penetration without adding an actuator for each new
function? Answering it requires reconciling three load regimes that
would ordinarily call for separate drives: high-speed, low-torque rotor
operation for lift; a brief, high-reaction arm deployment after which
the arms must carry the vehicle's weight; and sustained, high-torque,
low-speed rotation for drilling.

\textbf{Design requirements.} The task yields five requirements.
\emph{R1, flight-compatible mass:} planting hardware must fit within the
payload margin of a compact quadrotor, so new functions should reuse
existing actuators or structure where possible.
\emph{R2, load-bearing stance:} during drilling, reaction torque and
downward force must be carried by ground contacts rather than by thrust,
and the same stance must support rolling between locations.
\emph{R3, controllable tool reach:} the drill must reach the required depth
while keeping the chassis, battery, and drivetrain clear of the ground.
With \SI{127}{\milli\metre} propeller-guard wheels, the effective
ground-clearance radius is about \SI{65}{\milli\metre}; reserving roughly
\SI{30}{\milli\metre} as structural margin leaves about
\SI{35}{\milli\metre} of vertical travel, which the tray stroke and
drill-bit geometry conservatively realize as a maximum drilling depth of
\SI{30}{\milli\metre}, within which the drilling depth is controlled. This
range covers the shallow-burial depths above, although final seed depth
also depends on how the seed settles in the borehole.
\emph{R4, unpowered holding:} both terminal configurations must hold
without continuous current, so that maintaining a pose does not drain
the flight battery.
\emph{R5, isolation of flight-critical control:} mechanism commands must
not override attitude stabilization, and drilling and thrust must never
be demanded of the propulsion motors at the same time.
Sections~\ref{sec:system}--\ref{sec:control} address R1--R5.

\section{System Overview}
\label{sec:system}

The colour coding used in the morphing sequence of
Fig.~\ref{fig:configs} identifies six functional subsystems:
\textcolor{red}{\rule{6pt}{6pt}}~propellers and propeller-guard rings
(which double as the rover's ground-contact points);
\textcolor[gray]{0.85}{\rule{6pt}{6pt}}~primary morphing mechanism
(electronically-coupled dual-motor 1000:1 geared DC assembly);
\textcolor[rgb]{0,0.55,0}{\rule{6pt}{6pt}}~seed hopper;
\textcolor[rgb]{0.85,0.75,0}{\rule{6pt}{6pt}}~front rack-and-pinion
assembly (\SI{10}{\milli\metre} descent stroke);
\textcolor[rgb]{0.1,0.3,0.8}{\rule{6pt}{6pt}}~seed-gate tray;
\textcolor{black}{\rule{6pt}{6pt}}~gear train and drill bits.

The architecture uses an electronically coupled dual-motor drive to fold
the rear arms into ground contacts, a compact front descent stage to
position the drill, and the propulsion motors to drive the drill in plant
mode (Fig.~\ref{fig:configs}).

The vehicle operates as a conventional multirotor in flight mode (arms
extended, front tray raised, drill retracted) and as a four-point ground
rover in rover/plant mode (rear arms grounded on the propeller guards,
front tray descended, drill engaged). The mission loop is:
\begin{enumerate}
\item Fly to a target location using the autopilot.
\item On ground contact, morph to rover configuration: the rear arms
      fold outward onto the propeller-guard rings, grounding the
      vehicle. The front tray is then lowered to engage the drill gear
      train with the brushless motor shafts.
\item Drill into the soil for a configurable duration
      ($t_\mathrm{drill}\in[\SI{5}{\second},\SI{10}{\second}]$) to form
      the planting hole.
\item Retract the front tray and cycle the seed gate to dispense a seed
      into the hole.
\item Either advance on the propeller guards to the next site in the
      planting row and repeat steps 3--4, or, if the planting task is
      complete, morph back to flight configuration and fly to the next
      waypoint.
\end{enumerate}
Flight control is handled entirely by a dedicated flight controller
running the open-source ArduPilot stack, with missions defined in Mission
Planner~\cite{ardupilotmissionplanner}. All morphing logic runs on a
decoupled companion microcontroller, keeping flight-critical code
isolated from morphing logic, a deliberate architectural separation
motivated by the single-point-of-failure analysis in similar hybrid
platforms~\cite{sihite2023m4,yeh2025verticallyfoldable}.

\section{Actuator-Minimal Morphing Mechanism}
\label{sec:mechanism}

\subsection{Design Rationale: Actuator Reuse Across Flight, Locomotion,
  and Drilling}
\label{sec:rationale}

Table~\ref{tab:comparison} summarizes the actuation arrangements of
representative platforms. The present design combines a coordinated
rear-arm motor pair with a front-tray servo stage and reuses the
propulsion motors for drilling. Two mechanical design decisions support
this arrangement.

First, the rear-arm drive uses a shared control channel to command two
mechanically independent gear trains. This coordinates their inputs but
does not guarantee equal arm motion under unequal loads. The mechanical
end-stops define the terminal configurations, and the non-backdrivable
gearing holds the deployed arms as load-bearing ground contacts without
continuous motor current.

Second, the front tray acts as a mechanical clutch. Its
\SI{10}{\milli\metre} descent meshes the propulsion-motor shafts with the
drill gear train, allowing the flight motors to supply drilling torque.
Separate flight and drill commands prevent simultaneous thrust and drill
operation. The same tray motion can be repeated to tamp a seed that does
not settle fully into the borehole.

\subsection{Primary Drive and Kinematic Coupling}
\label{sec:kinematics}

The primary morphing actuator is an electronically-coupled dual-motor
assembly: two independent geared DC motors (\SI{12}{\volt}, \SI{30}{RPM},
1000:1 reduction, non-backdrivable spur gearbox), one per rear folding
arm, wired in parallel and driven from a single control channel.
Let $d_i \in [0,\,d_\mathrm{max}]$ denote the rack displacement of arm
$i \in \{\mathrm{L},\mathrm{R}\}$, with $d_i=0$ in flight mode and
$d_i=d_\mathrm{max}=\SI{30}{\milli\metre}$ in plant mode. A nominal
linear model relates each arm's deployment angle to its own rack travel:
\begin{equation}
  \theta_i(d_i) \;\simeq\; \theta_{\max,i}\,\frac{d_i}{d_\mathrm{max}},
  \quad i \in \{\mathrm{L},\mathrm{R}\}.
  \label{eq:arm_angle}
\end{equation}
Here $\theta_{\max,i}$ is the terminal deployment angle of each arm.
The shared drive signal coordinates the motor commands; it does not
constrain $d_\mathrm{L}=d_\mathrm{R}$ or
$\theta_\mathrm{L}=\theta_\mathrm{R}$. Differences in friction and load
can produce unequal motion, and synchronization under variable loading
is not quantified here. At the deployment end-stops, the arms serve as
rear ground contacts. The non-backdrivable gearboxes retain these
terminal configurations without holding current.

\subsection{Front Tray and Drill Engagement}
\label{sec:secondary}

The front motor tray is actuated by a dedicated servo-driven
rack-and-pinion (\SI{10}{\milli\metre} stroke) and is commanded
independently of the primary motor assembly:

\textbf{Drill engagement.}  Two mirrored servos
(2.2\,kg{\textperiodcentered}cm torque each) drive the front tray
downward by \SI{10}{\milli\metre}.  As the tray descends, the
brushless propulsion motor shafts engage a drill-driving gear train,
repurposing the flight motors as drill actuators and eliminating a
dedicated drill drive.  Maximum drilling depth is approximately
\SI{30}{\milli\metre}, set by the tray stroke and drill-bit geometry
within the ground-clearance budget the propeller-guard wheels afford.

\textbf{Seed gate.}  A dedicated geared DC motor slides open a
rack-and-pinion gate cover at the base of the seed hopper once the drill
has retracted and the outlet is aligned with the borehole, dispensing
a seed into the cavity before closing again. The seed-box floor slopes
forward by approximately $15^{\circ}$, allowing gravity to feed the remaining seeds
toward the dispensing outlet as the hopper empties.

\section{Morphing and Seed-Gate Control}
\label{sec:control}

Control spans two layers.
Airborne, attitude and rate stabilization are delegated in full to the
onboard autopilot, a commercial flight controller (Table~\ref{tab:platform})
running the open-source ArduPilot stack~\cite{ardupilotmissionplanner},
whose cascaded PID rate-and-attitude loops close on the integrated IMU
and compass at high rate, the well-established quadrotor control
structure of~\cite{bouabdallah2004pidlq}. We retune these gains for the
platform's shifted post-morph mass distribution and run them on hardware
separate from the reconfiguration logic, so flight-critical control is
never overridden by a morph command. ArduPilot and Mission Planner
provide IMU- and compass-based stabilization and support GNSS waypoint
missions~\cite{ardupilotmissioncommands}. The companion microcontroller
accepts separate commands for the rear morph drive, front tray, and seed
gate. Automatic triggering of those commands from autopilot landing and
mission events remains to be integrated.

\subsection{Timing-Based Morph Sequence}
\label{sec:timing}

\begin{figure}[!t]
  \centering
  \includegraphics[width=\columnwidth]{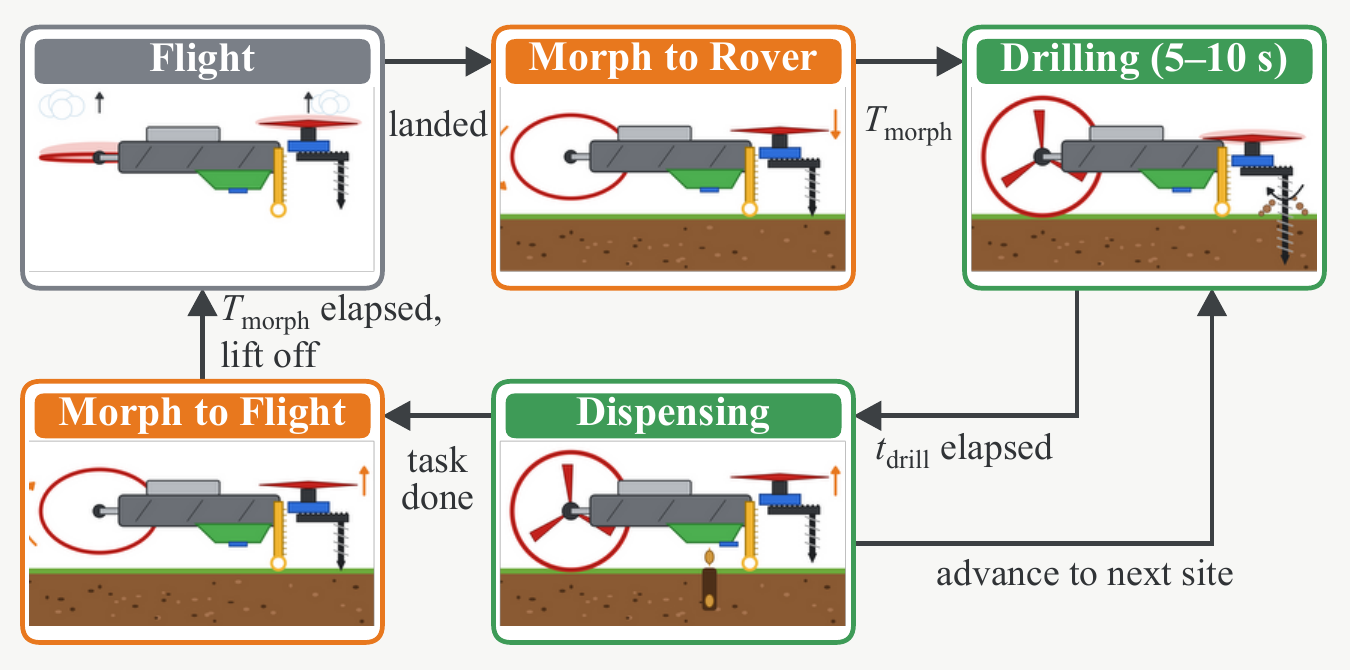}
  \caption{Mission-level morphing and planting sequence, with the
    platform drawn in each state (colours as in Fig.~\ref{fig:configs}).
    Morph transitions use a fixed drive interval; mechanical end-stops
    define the terminal configurations. Grey: airborne. Orange: morph
    transition. Green: grounded task phases. The
    row-advance branch repeats \textsc{Drilling}$\to$\textsc{Dispensing}
    at successive sites without re-entering flight.}
  \label{fig:fsm}
\end{figure}

Figure~\ref{fig:fsm} organizes the operating sequence into five mission
states: \textsc{Flight}, \textsc{MorphToRover}, \textsc{Drilling},
\textsc{Dispensing}, and \textsc{MorphToFlight}. In the present
prototype, a companion controller exposes separate commands for the rear
morph drive, front tray, and seed gate. The sequence is executed by
issuing these commands in the order shown; integration with the
autopilot's landing and waypoint events is future work.

The rear-arm drive is open loop. For direction
$\sigma\in\{-1,+1\}$, the implemented command is
\begin{equation}
 u_m(t)=
 \begin{cases}
   \sigma, & 0\leq t<T_d,\\
   0, & t\geq T_d,
 \end{cases}
 \qquad T_d=\SI{1.65}{\second}.
 \label{eq:morph_command}
\end{equation}
The fixed interval is selected to carry both rear arms to their
mechanical end-stops. Arm position is not measured during the transition,
so endpoint accuracy depends on reaching those stops within $T_d$.

Position-commanded actuators follow a cubic ease-in/ease-out profile
between their end angles rather than a step command:
\begin{equation}
  \theta_k(t) = \theta_{k,0} + \Delta\theta_k\,(3\tau^2 - 2\tau^3),
  \qquad \tau = t/T_k \in [0,1],
  \label{eq:scurve}
\end{equation}
where $T_k$ is the transition duration of actuator $k$. The profile
starts and ends with zero velocity, avoiding the shock of a step
command, and is realized in firmware as a fine incremental
interpolation between end angles.

\subsection{State Persistence}
\label{sec:compensation}

The left and right rear-arm motors are mechanically independent but
parallel-driven from a single control channel and share one drive signal. Manufacturing tolerances in the PETG-printed gearboxes can still
produce timing differences between the two sides during the transition;
it is the mechanical end-stops, not the shared drive signal, that bring
each arm to its terminal configuration, provided the allotted drive
time is sufficient to reach the stop. On every state
transition the companion microcontroller writes the corresponding binary
morph or tray state to a reserved EEPROM address, so that on power-up the firmware
restores the last stored logical state. This does not verify physical
pose: recovery after an interrupted transition requires checking the
mechanism configuration before resuming operation.

\subsection{Seed-Gate Controller}
\label{sec:seedgate}

\begin{figure}[!t]
  \centering
  \includegraphics[width=\columnwidth]{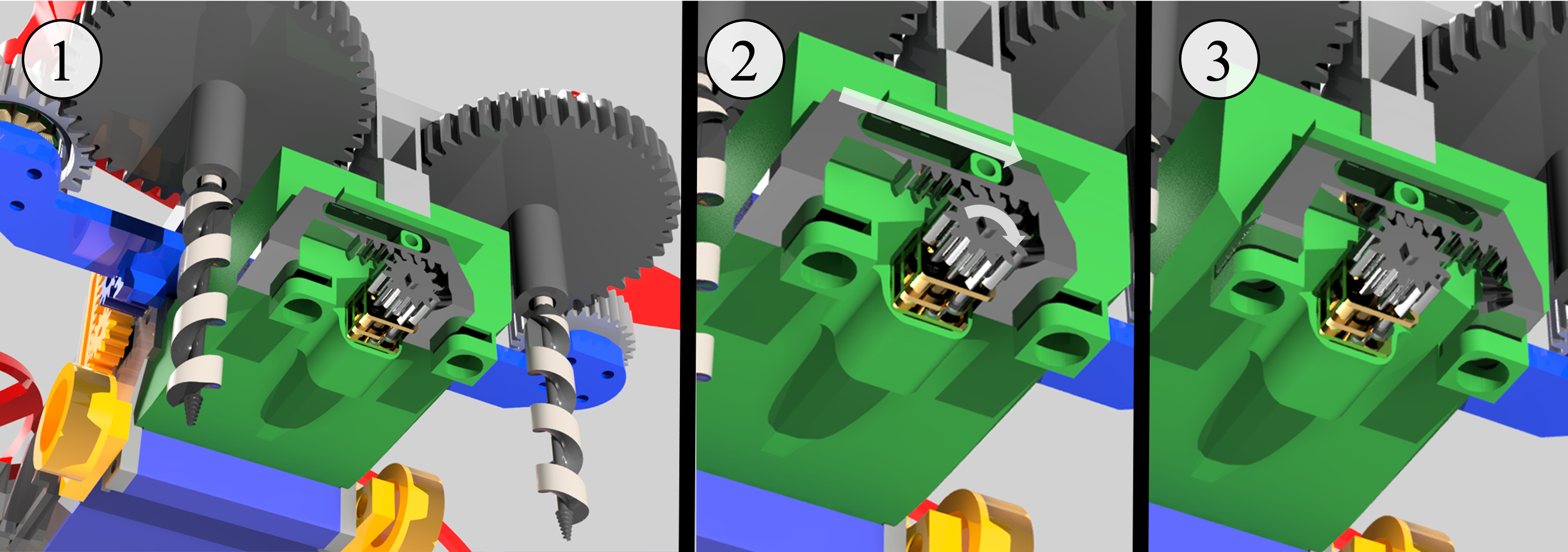}
  \caption{Seed-gate dispensing mechanism. \textbf{(1)}~Underside
    view: seed box, rack-and-pinion gate, and twin drill augers.
    \textbf{(2)}~\emph{Open}: the gate motor rotates the pinion,
    sliding the cover aside to release seed.
    \textbf{(3)}~\emph{Closed}: from (2) the gear rotates clockwise, sealing the box.
    The command uses an 80\,ms opening pulse, a 1\,s dwell, and an
    80\,ms closing pulse.}
  \label{fig:seedgate}
\end{figure}

The seed gate uses a sliding cover driven by a dedicated geared DC
motor via a rack-and-pinion mechanism (Fig.~\ref{fig:seedgate}), and
its cover features two dispensing slots per actuation cycle. The
implemented cycle applies an \SI{80}{\milli\second} opening command,
a \SI{1}{\second} dwell, and an \SI{80}{\milli\second} closing command,
for a nominal duration of \SI{1.16}{\second}. The mechanism supports
seeds up to \SI{10}{\milli\metre}
in diameter; the PETG-printed gate insert can be swapped for larger or
smaller seed profiles without structural changes. The dispensing command
is issued after rear-arm deployment and drilling are complete.

\section{Experimental Results}
\label{sec:experiments}

\subsection{Drill-and-Seed Cycle}
\label{sec:drillseed}

\begin{figure*}[!t]
  \centering
  \includegraphics[width=\textwidth]{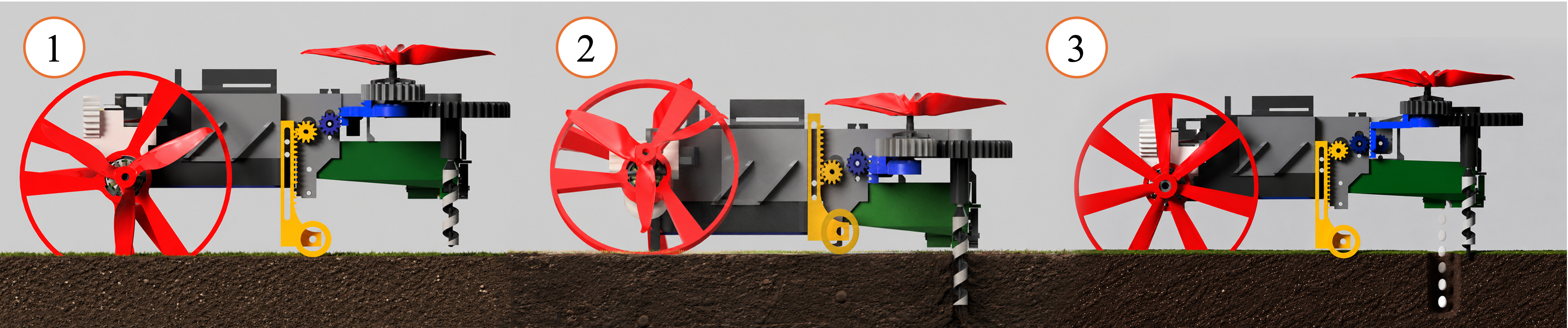}
  \caption{Drill-and-seed cycle (side-view cross-section, soil shown
    below the ground line).
    \textbf{(1)}~Plant mode with front tray raised; drill bits above soil.
    \textbf{(2)}~Front tray descends via rack-and-pinion; propulsion motors
    re-engage the drill gear train, driving the bits into the substrate.
    \textbf{(3)}~Tray retracts; the vehicle advances so the seed-gate
    outlet sits above the borehole, then dispenses a seed.
    If the seed does not settle into the hole, step~(2) is repeated:
    the tray re-descends and the drill tip tamps the seed in before
    final retraction.}
  \label{fig:drilling}
\end{figure*}

Fig.~\ref{fig:drilling} illustrates the complete drill-and-seed cycle
from a side-view cross-section of the platform above the soil.
In panel~\textbf{1} the vehicle is in the fully grounded plant
configuration (Stage~5 of Fig.~\ref{fig:configs}): all four
propeller-guard wheels are on the ground, the front tray is in its
raised position, and the drill bits are held above the soil surface.
In panel~\textbf{2} the rack-and-pinion lowers the front tray; the
front brushless motors spin up and their shafts mesh with the drill-drive
gear train, driving the bits into the substrate. The propellers are
deliberately re-engaged at this stage so that the motor shafts couple
with the drill gears via the descended tray: the same motors that
provide lift in flight thus provide torque for drilling, eliminating a
dedicated drill actuator. In panel~\textbf{3} the tray retracts and the
vehicle advances a short distance so the seed-gate outlet sits directly
above the borehole, then dispenses a seed. Whether the seed settles
cleanly into the hole depends on terrain: on loose or uneven substrate
it can fail to enter on the first attempt, in which case the motion of
panel~\textbf{2} is repeated: the tray re-lowers and the drill tip tamps
the seed into the hole before final retraction. The robot is then back in
the grounded configuration of panel~\textbf{1}, ready either to advance
to the next planting site on its deployed rear wheels or to morph back
to flight.

\subsection{Experimental Platform}
\label{sec:platform}

\textbf{The airframe is 3D-printed in PETG for rapid iteration, with carbon-fibre tube
arms carrying the structural load; the propeller-guard rings double as
the passive ground-rolling wheels once the rear arms deploy, so no
separate wheel component is needed.} Table~\ref{tab:platform}
lists the propulsion, autopilot, morphing, and seed-handling components
used in the prototype.

\begin{table}[!t]
  \centering
  \caption{Platform Hardware Specifications}
  \label{tab:platform}
  \scriptsize
  \setlength{\tabcolsep}{2.5pt}
  \begin{tabular}{@{}p{2.35cm} p{2.45cm} p{3.35cm}@{}}
    \toprule
    Parameter & Value & Notes \\
    \midrule
    Airframe mass               & $<$\SI{500}{\gram} & battery excluded \\
    Overall size                & $286.83\times126.1$\,mm & flight mode \\
    Wheel diameter              & \SI{127}{\milli\metre} & 5\,in propeller guard \\
    Battery capacity            & 4S, 3500\,mAh LiPo & all-up mass $\sim$800\,g \\
    Static thrust               & 5.3\,kgf total & 4S; thrust-to-weight $>$6:1 \\
    \midrule
    Brushless motors            & Emax ECOII-2306, 1700\,KV & $\times$4 \\
    Propeller                   & 5054 & $\times$4 \\
    ESC                         & Cyclone 45\,A BLHeli\_S & $\times$4 \\
    Flight controller           & Pixhawk 2.4.8 & ArduPilot firmware; Mission Planner GCS \\
    Companion MCU               & Espressif ESP32-S3 & morphing + seed commands; EEPROM state \\
    \midrule
    Morph motors                & 2$\times$ geared DC, 30\,RPM, 1000:1 & 12\,V; coupled command \\
    Rack travel                 & \SI{30}{\milli\metre} & rear-arm deployment \\
    Tray actuators              & Servo, 2.2\,kg{\textperiodcentered}cm & $\times$2 mirrored \\
    Tray stroke                 & \SI{10}{\milli\metre} & drill engagement \\
    Drill depth                 & $\leq$\SI{30}{\milli\metre} & controlled \\
    Morph interval              & \SI{1.65}{\second} & fixed drive command \\
    Gate cycle                  & \SI{1.16}{\second} & 80\,ms open + 1\,s dwell + 80\,ms close \\
    Max seed diameter           & \SI{10}{\milli\metre} & swappable insert \\
    \bottomrule
  \end{tabular}
\end{table}

\subsection{Performance}
\label{sec:performance}

The motors produce a scale-measured static thrust of \SI{5.3}{kgf} on 4S,
a thrust-to-weight ratio above 6:1 at the \SI{0.8}{\kilo\gram} all-up mass.
Flight tests with the 4S, 3500\,mAh battery gave a hover endurance of
approximately \SI{10}{\minute}. In rover configuration, the vehicle
drives on its \SI{127}{\milli\metre} propeller-guard wheels. A nominal
wheel speed $n=\SI{100}{RPM}$ gives the kinematic estimate
\begin{equation}
 v_g=\frac{\pi Dn}{60}=\SI{0.67}{\metre\per\second},
 \label{eq:ground_speed}
\end{equation}
before slip and drivetrain losses; ground speed was not measured in
these trials. Drilling depth is controlled up to approximately
\SI{30}{\milli\metre}, and the full
drill-and-seed cycle (tray descent, drilling, retraction, and
dispensing) completes in approximately \SI{20}{\second} per site.

\textbf{Mission-energy sensitivity.} To quantify the possible benefit
of replacing short inter-site flights with ground motion, consider an
illustrative duty-cycle model. It was experimentally found that the  traversal and drilling power be $0.2$
and $0.4$ times average flight power, respectively. The time-weighted
ground power ratio is
$r_g=0.2w_r+0.4w_d$, where $w_r+w_d=1$ are ground-time fractions.
Using the observed \SI{10}{\minute} hover baseline, the corresponding
mission duration is
\begin{equation}
  T_{\mathrm{mission}} = t_f + \frac{T_{\mathrm{air}}-t_f}{r_g},
  \qquad T_{\mathrm{air}}=\SI{10}{\minute},
  \label{eq:energy_runtime}
\end{equation}
where $t_f$ includes outbound, inter-site, and return flight.
Equal traversal and drilling times give $r_g=0.3$: with
$t_f=\SI{2}{\minute}$, runtime is \SI{28.7}{\minute}, versus a
\SI{33.3}{\minute} ground-only limit. The corresponding ground-operation
runtime multiplier relative to continuous flight is
$G_g=1/r_g=3.33$.

\begin{figure}[!t]
  \centering
  \includegraphics[width=\columnwidth]{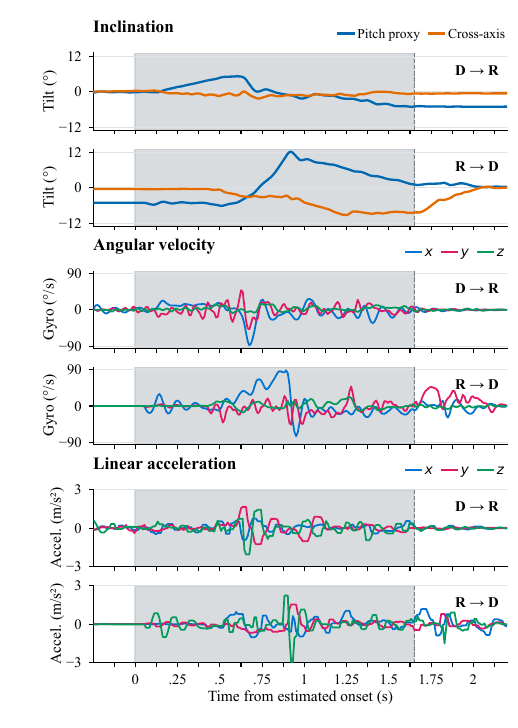}
  \caption{Paired morphing responses: drone-to-rover (D$\to$R) above
    rover-to-drone (R$\to$D) for each signal. Tilt is referenced to the
    initial stationary median (0.5--1.5\,s). Gray shading indicates the
    nominal 1.65\,s drive interval aligned to estimated onset.}
  \label{fig:morph_imu}
\end{figure}

Figure~\ref{fig:morph_imu} compares one continuous drone--rover--drone
cycle recorded by a smartphone IMU at approximately \SI{101}{\hertz}.
The pitch proxy settles $5.07^{\circ}$ from its initial reference in rover
mode; the reverse transition has a larger transient before returning near
the initial inclination. Sensor-$x$ tilt is used as a pitch proxy based on
the reported fore--aft motion. The current firmware specifies a nominal
\SI{1650}{\milli\second} drive interval, shown from estimated onsets at
3.2 and 11.3\,s in the recording. This alignment supports comparison of
the two responses; it is not a measurement of command-to-completion time.

During drone-to-rover reconfiguration, the pitch proxy first rises and
then rapidly levels, accompanied by a pronounced gyroscope and
acceleration transient approximately 0.65--0.70\,s after estimated onset.
This response is consistent with the deployment geometry: the front
passive wheels descend from the start, whereas the rear propeller guards
begin supporting the chassis later, after roughly $40^{\circ}$ of
rear-arm rotation. Rear-wheel ground contact changes the support
configuration and provides a mechanical explanation for the transient
and subsequent leveling. The resulting fore--aft motion may also assist
the inclined hopper floor in moving seeds retained at the rear toward
the dispensing cavity. These contact and seed-transport effects are
mechanical interpretations of the inertial response, rather than direct
measurements of arm angle or seed motion.

Let $\widetilde{\theta}_{x,0}$ and $\widetilde{\theta}_{y,0}$ denote
the median stationary angles over 0.5--1.5\,s. The plotted inclination
changes are computed from the gravity vector as
\begin{equation}
\begin{aligned}
 \Delta\theta_x(t)&=\frac{180}{\pi}\operatorname{atan2}(g_y,g_z)
                     -\widetilde{\theta}_{x,0},\\
 \Delta\theta_y(t)&=\frac{180}{\pi}\operatorname{atan2}\!\left(
                     -g_x,\sqrt{g_y^2+g_z^2}\right)-\widetilde{\theta}_{y,0},
\end{aligned}
\label{eq:imu_tilt}
\end{equation}

\subsection{Morph Repeatability and Substrate Dependence}
\label{sec:repeatability}

Five consecutive flight-to-plant-to-flight cycles on a rigid bench
completed deployment and retraction with the \SI{1.65}{\second} drive
interval. This short sequence verifies repeated endpoint reach under the
tested support condition; a larger trial with arm-angle and motor-current
measurements is needed to quantify timing dispersion under load.

\subsection{Seed Placement Trials}
\label{sec:seeding}

Across 50 drill-and-dispense cycles performed outdoors on natural
ground under ambient conditions (tray descent, drilling, retraction, one
seed-gate actuation), the seed entered the borehole in 47 trials. The
observed success fraction is $\hat p=47/50=0.94$, with a 95\,\% Wilson
interval of $[0.838,\,0.979]$. Success denotes entry into the borehole
after one gate actuation; germination was not evaluated in this trial.
The three failures were one deformed seed that missed the gate
alignment, one seed that landed outside the borehole, and one seed
displaced by wind after release.

\section{Discussion and Limitations}
\label{sec:discussion}

This work evaluates a proof-of-concept platform through bench morphing
tests and outdoor drill-and-dispense trials. These results demonstrate
the reconfiguration and seed-placement mechanisms; field deployment
requires broader terrain testing and integration of the mission stages.

\textbf{Open-loop timing under load.} The timing-based controller
(Section~\ref{sec:timing}) was calibrated on a rigid bench surface.
Variable soil resistance during rear-arm deployment could desynchronize
the timed sequence from the true mechanical state; a current-sensing
stall detector on the primary motor assembly is a natural closed-loop
extension that bounds worst-case timing error while preserving the
geometry-assisted design.

\textbf{Navigation and mission integration.} The prototype uses
ArduPilot and Mission Planner for stabilization and waypoint flight
(Section~\ref{sec:control}), while morphing and planting are issued as
separate companion-controller commands. Triggering those commands from
autopilot landing and mission events is the next integration step.
As the hopper empties, the changing seed mass alters the centre-of-gravity
shift associated with rear-arm reconfiguration. Its influence on attitude
estimation after returning to flight warrants characterization, consistent
with the mass-distribution considerations in other morphing
platforms~\cite{falanga2019foldable,fabris2020geometry}.

\textbf{Environmental robustness.} No controlled wind or disturbance
testing has been performed, although one outdoor placement failure was
caused by wind. Front-tray descent and drill engagement assume a roughly
level landing attitude; performance on sloped or uneven terrain, which
is common at restoration sites, remains to be evaluated.

\textbf{Terrain-aware deployment.} A future LiDAR or depth-camera module
could map local slope, obstacles, and available ground clearance to guide
landing-site selection and the decision to reconfigure. Combined with
arm-position and contact sensing, these measurements could adapt the
deployment sequence and drive duration to local terrain. The shared
rear-arm command would still coordinate both sides; independently
controlled terrain conformation would require additional actuation or
mechanical compliance.

\textbf{Seed germination.} The platform is motivated by the survival
advantage of embedded over surface-placed
seed~\cite{luo2023selfburying,joyce2024predation}; this work establishes
the embedding \emph{capability}, but a direct comparison of germination
and establishment against broadcast or seed-ball methods requires a
separate field trial.

\textbf{Mechanical durability.} The PETG gear train, rack-and-pinion,
and drill coupling enable rapid prototyping. Repeated loaded-cycle tests
would quantify wear, backlash, and changes in deployment timing. Metal
gear interfaces and stiffer structural components are candidates for
improving service life; their benefit should be evaluated together with
the added mass and changes in flight dynamics.

\section{Conclusion}
\label{sec:conclusion}

We presented a proof-of-concept aerial-ground robot for flight, ground
repositioning, and controlled-depth seed embedding. \textbf{The core novelty is the
hardware}: a geometry-assisted, tri-functional morphing robot, with
an electronically coordinated dual-motor
drive, realized and validated as a physical prototype. An integrated
control system sequences flight, morphing, and seed dispensing with
geometry-assisted timing and state persistence across power cycles, and
demonstrated ground locomotion on the deployed rear wheels enables
multi-site planting in a single sortie.

LiDAR-guided deployment would extend this architecture toward
terrain-adaptive planting missions.
 
\bibliographystyle{IEEEtran}
\bibliography{references2}

\end{document}